\pdfoutput=1

\documentclass[11pt]{article}

\usepackage{ACL2023}

\usepackage{times}
\usepackage{latexsym}

\usepackage[T1]{fontenc}

\usepackage[utf8]{inputenc}

\usepackage{microtype}

\usepackage{inconsolata}

\usepackage{graphicx}
\usepackage{booktabs}
\usepackage{multirow}
\usepackage{float}
\usepackage{todonotes}
\usepackage{siunitx}
\usepackage{array,multirow,graphicx}
\newcommand{\turna}{\textsc{Turna}}
\newcommand{\turnaenc}{\turna-Encoder}
\newcommand{\berturk}{\textsc{bert}urk}
\newcommand{\bert}{\textsc{bert}}
\newcommand{\convberturk}{\textsc{C}onv\bert urk}

\newcommand*\rot{\rotatebox{90}}

\newcommand\turnalogo{\raisebox{-2pt}{\includegraphics[width=0.9em]{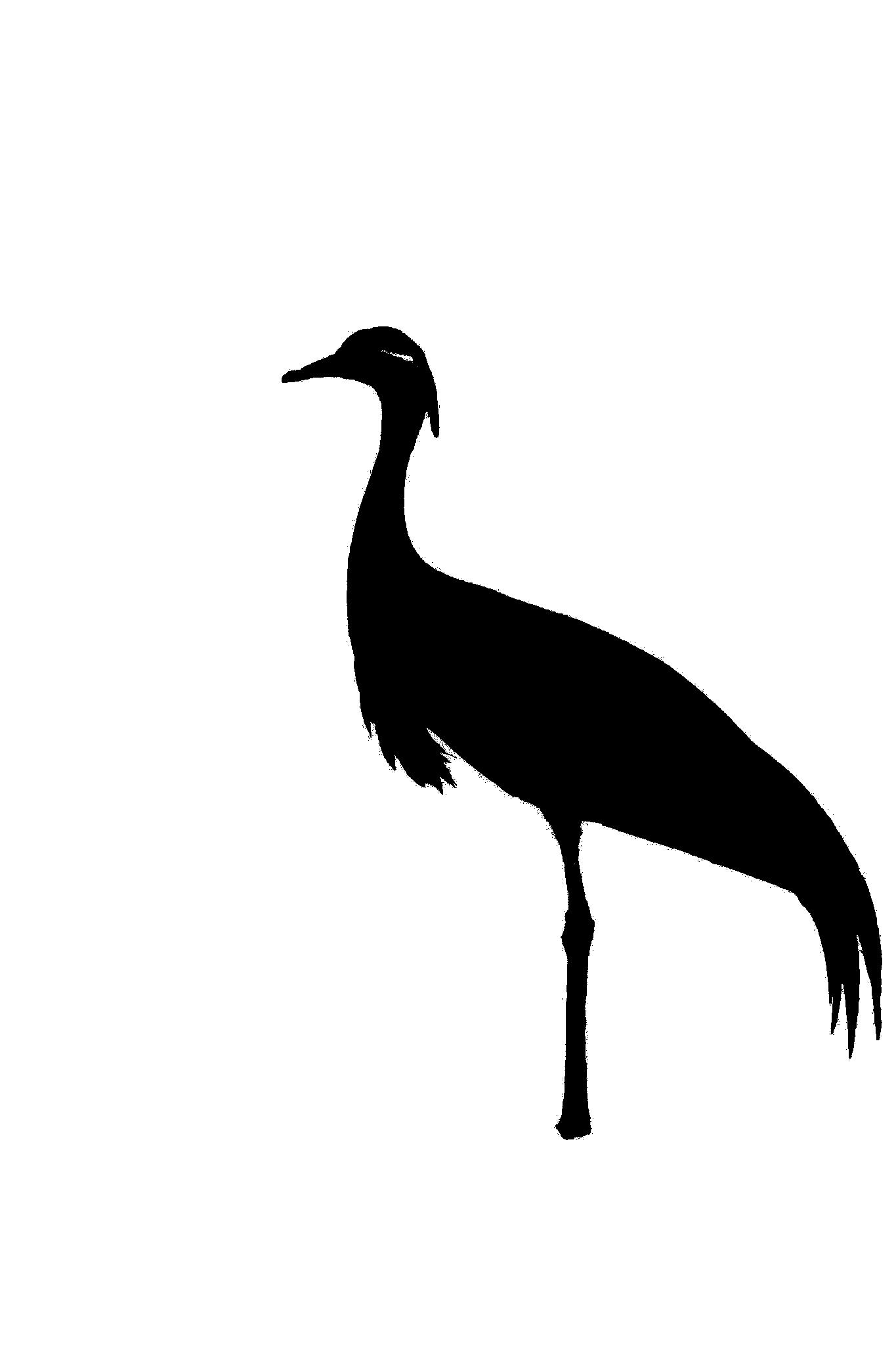}}}

\title{\turnalogo\ \turna: A Turkish Encoder-Decoder Language Model for Enhanced Understanding and Generation}

\author{Gökçe Uludoğan$^1$  \and Zeynep Yirmibeşoğlu Balal$^1$ \and Furkan Akkurt$^1$  \\ 
       {\bf Melikşah Türker$^{1,2}$} \and {\bf Onur Güngör$^1$} \and {\bf Susan Üsküdarlı$^1$} \\
         $^1$Department of Computer Engineering, 
         Bogazici University \\
         $^2$  VNGRS-AI\\
         \texttt{\{gokce.uludogan,furkan.akkurt,zeynep.yirmibesoglu,} \\ \texttt{meliksah.turker,onurgu,suzan.uskudarli\}@bogazici.edu.tr}}

\begin{document}


\newpage
\maketitle

\begin{abstract}
The recent advances in natural language processing have predominantly favored well-resourced English-centric models, resulting in a significant gap with low-resource languages.
In this work, we introduce the language model \turna, which is developed for the low-resource language Turkish and is capable of both natural language understanding and generation tasks.
\turna\ is pretrained with an encoder-decoder architecture based on the unified framework UL2 with a diverse corpus that we specifically curated for this purpose. 
We evaluated \turna\ with three generation tasks and five understanding tasks for Turkish.
The results show that \turna\ outperforms several multilingual models in both understanding and generation tasks, and competes with monolingual Turkish models in understanding tasks.
\turna\ is made available at \texttt{\href{https://huggingface.co/boun-tabi-LMG/turna}{hf.co/boun-tabi-LMG/turna}}.
\end{abstract}

\section{Introduction}

Recent advances in natural language processing (NLP) have predominantly produced English-centric models~\cite{devlin-etal-2019-bert,DBLP:conf/iclr/ClarkLLM20, radford2019language,NEURIPS2020_1457c0d6,touvron2023llama,jiang2023mistral}.
English-centric models have benefited from the vast amount of training data gathered from an abundance of English resources present on the web.
As such, these models become utilized in applications and fuel an abundance of further research leading to the state-of-the-art performances across various tasks~\cite{touvron2023llama,jiang2023mistral}.
On the other hand, low-resource languages face challenges due to the lack of data and limited computational resources, leading to a significant gap between models trained on well-resourced languages and those focusing on low-resource languages.
Multilingual models have been proposed that attempt to bridge this gap~\cite{devlin-etal-2019-bert, conneau-etal-2020-unsupervised, xue-etal-2021-mt5,liu-etal-2020-multilingual-denoising}. 
However, they often do not perform well in tasks requiring a deep understanding of language-specific nuances, such as dependency parsing and named entity recognition~\cite{virtanen2019multilingual,baumann-2019-multilingual,tanvir-etal-2021-estbert} and lag behind monolingual models of the same scale~\cite{rust-etal-2021-good,nozza2020mask}.

Recently, pretrained language models built upon transformers~\cite{vaswani2017attention} have dominated NLP. 
These models vary in terms of their architectures and objectives (i.e., causal language modeling and denoising objectives). 
The architectures are commonly classified as encoder-only, decoder-only, or encoder-decoder models.
Encoder-only models are typically trained with denoising objectives and 
 focus on understanding tasks~\cite{devlin-etal-2019-bert,DBLP:conf/iclr/ClarkLLM20}.
Decoder-only models are designed for generation tasks with causal language modeling~\cite{radford2019language,NEURIPS2020_1457c0d6,touvron2023llama}.
Finally, encoder-decoder models deal with NLP tasks that require both understanding and generation of texts~\cite{dong2019unified,tay2023unifying}. 
Towards this end, the Text-to-Text Transformer (T5)~\cite{raffel2020exploring} employs an encoder-decoder architecture that was pretrained with a denoising objective known as span corruption. 
The Unifying Language Learning (UL2) framework ~\cite{tay2023unifying} proposes the Mixture-of-Denoisers (MoD) pretraining objective that combines several denoising objectives. 
By coupling the MoD objective with an encoder-decoder architecture, they have achieved state-of-the-art results in a range of NLP tasks.

For Turkish, while encoder-only models exist~\cite{stefan_schweter_2020_3770924}, there is a need for large-scale pretrained models that can perform both natural language understanding (NLU) and generation (NLG).
This work aims to develop such a model for Turkish that performs well across a variety of tasks. 
Towards this end, we first compile a diverse range of corpora for pretraining, including web data, scientific articles, graduate theses, books, creative writing, and transcriptions of parliamentary speech. Subsequently, we pretrain \turna,
which adopts an encoder-decoder architecture following the UL2 framework~\cite{tay2023unifying}. 
Furthermore, we collect several publicly available Turkish datasets from a variety of tasks for benchmarking purposes. 
Our contributions are as follows:
\begin{itemize}
    \item The release of  \turna\footnote{\texttt{\href{https://huggingface.co/boun-tabi-LMG/turna}{hf.co/boun-tabi-LMG/turna}}}, 
      the first unified language model capable of both understanding and generation tasks in Turkish. 
      Thus far, this model is the largest of its kind, which has 1.1B parameters and is trained on a diverse range of corpora consisting of ~43B tokens from various domains.
    \item The evaluation of \turna\ on 13 datasets across eight tasks, showing that it surpasses multilingual models across many tasks and it either outperforms or is on par with BERTurk~\cite{stefan_schweter_2020_3770924} (a Turkish monolingual encoder-only model) in understanding tasks.
    \item The release of a  publicly available open source code for data collection, filtering\footnote{\texttt{\href{https://github.com/boun-tabi-LMG/turkish-academic-text-harvest}{github.com/boun-tabi-LMG/turkish-academic-\\text-harvest}}}, model training\footnote{\texttt{\href{https://github.com/boun-tabi-LMG/turna}{github.com/boun-tabi-LMG/turna}}}, and fine-tuning\footnote{\texttt{\href{https://github.com/boun-tabi-LMG/turkish-lm-tuner}{github.com/boun-tabi-LMG/turkish-lm-tuner}}}.

\end{itemize}

\section{Related Work}
\label{sec:related work}

\subsection{Multilingual Language Models}
Multilingual language models are used for tasks that involve multiple languages and resources with limited data. Turkish, considered as a low-resource language, is moderately represented in these models, such as mBERT~\cite{devlin-etal-2019-bert}, XLM-R~\cite{conneau-etal-2020-unsupervised}, mBART~\cite{liu-etal-2020-multilingual-denoising}, mT5~\cite{xue-etal-2021-mt5}, XGLM~\cite{lin-etal-2022-shot}, mGPT~\cite{shliazhko2022mgpt}, and mDeBERTa~\cite{he2023debertav3}. 
However, these models' scale-to-performance ratios are often not as efficient as that of monolingual models for language-specific applications with abundant data~\cite{rust-etal-2021-good,nozza2020mask}.

\subsection{Turkish Language Models}

A series of \bert\ models for Turkish known as \berturk\ have already been trained~\cite{stefan_schweter_2020_3770924}. This collection includes several variations of \bert~\cite{devlin-etal-2019-bert}, Distil\bert~\cite{sanh2019distilbert}, \convberturk ~\cite{jiang2020convbert}, and \textsc{electra}~\cite{DBLP:conf/iclr/ClarkLLM20}, with varying cases, vocabulary, and model sizes. 
Most models were trained on a 35GB corpus with 4.4B tokens, drawn from 
the Turkish \textsc{oscar} corpus~\cite{oscar_2022arXiv220106642A}, 
a Wikipedia dump, 
and various OPUS corpora~\cite{TIEDEMANN12.463}.
Some models, like \convberturk\ and ELECTRA, were also trained on the Turkish portion of the mC4 corpus. The models have been evaluated on various downstream tasks, such as part-of-speech tagging, named entity recognition, and question answering. 
They often outperform their multilingual counterparts, 
mBERT~\cite{devlin-etal-2019-bert} 
and XLM-R~\cite{conneau-etal-2020-unsupervised}. 
However, these models are all encoder-only models, most suitable for language understanding tasks. There is a clear need for Turkish models that excel in both language understanding and generation.

\subsection{Pretraining objectives}

Recently, pretrained language models based on transformers have been dominant in the NLP field, exhibiting variations in both components and objectives. Models that exclusively employ encoders, typically trained with denoising objectives, are geared toward understanding tasks, as exemplified in works such as \citet{devlin-etal-2019-bert} and \citet{DBLP:conf/iclr/ClarkLLM20}. Conversely, models that exclusively use decoders are designed for generation tasks, employing causal language modeling, as demonstrated in various studies~\cite{radford2019language,NEURIPS2020_1457c0d6,touvron2023llama}. The Text-to-Text Transformer (T5)~\cite{raffel2020exploring}, on the other hand, employs an encoder-decoder architecture and undergoes pretraining with a denoising objective referred to as span corruption. UniLM~\cite{dong2019unified} is also an encoder-decoder model, but pre-trained using unidirectional, bidirectional, and sequence-to-sequence language modeling. This can be seen as a combination of causal and denoising objectives. 
Recently, \citealp{tay2023unifying} proposed that various pretraining objectives can be recast as each other. They introduced the UL2 framework based on a pretraining objective called Mixture-of-Denoisers (MoD), which combines different pretraining paradigms. They compared decoder-only and encoder-decoder models trained with the MoD objective and found that encoder-decoder models often perform better. Notably, by using the MoD objective and moderately scaling up the model, they achieved state-of-the-art performance on a diverse set of NLP tasks including understanding and generation tasks. 

\section{Data}
\label{sec:data}

We compiled a diverse Turkish monolingual dataset to pretrain our model. Our dataset comprises of 
a web corpus, 
scientific corpora gathered from Turkish articles and graduate theses, 
Turkish books, 
a corpus of specially collected writings from Bilkent University, 
and transcriptions of parliamentary debates. 
The details of each corpus are explained in the following subsections, and the training corpora statistics are summarized in Table~\ref{tab:train_data}.

During data splitting, we ensured that the validation set of each dataset contain a minimum of \num{100}K tokens. The resulting train-validation splits are reported under each subsection. 

\subsection{Web Corpora}

mC4~\cite{raffel2020exploring} 
and OSCAR-2201~\cite{oscar_2022arXiv220106642A} 
are two large multilingual web corpora.
Their Turkish sections contain 87.7M and 10.8M web pages, respectively, and result in 98.5M web pages when combined.
Although significant in quantity, web data is full of noise that is not considered part of the natural language, such as titles and keywords.
Therefore, such corpora should be cleaned before being used for training.
The OSCAR and mC4 corpora used in this work were cleaned by the VNGRS-AI team using a set of heuristics and rules.
The cleaned version of the combined web corpus contains 50.3M pages.

\subsection{Scientific Corpora} \label{sec:sci_cor}

To create a corpus in the scientific domain characterized by its formal and informative language style, we collected articles and theses written in Turkish. We downloaded the articles from DergiPark\footnote{\texttt{\href{https://dergipark.org.tr}{dergipark.org.tr}}}, a major platform for Turkish academic journals. Our initial collection included \num{407146} articles, all in PDF format and labeled as Turkish. These articles were sourced from \num{1857} distinct journals, comprising a diverse range of topics. These articles form our \textit{Dergipark} scientific corpus.

In addition to articles, we also collected scientific texts in the form of theses. These theses, products of higher education in Turkey, were accessed from Turkey's National Thesis Center\footnote{\texttt{\href{https://tez.yok.gov.tr/UlusalTezMerkezi}{tez.yok.gov.tr/UlusalTezMerkezi}}}. From this repository, we downloaded \num{486166} theses marked as Turkish, which composes our \textit{YökTez} scientific corpus.  

The collected documents were in PDF format. For text extraction, we utilized the Apache Tika parser\footnote{\texttt{\href{https://github.com/apache/tika}{github.com/apache/tika}}}. 
We applied a rigorous cleaning and filtering strategy to remove undesired content like page numbers, equations, table entries, and similar unnecessary tokens introduced by the extraction process, as detailed in Section \ref{sec:filtering}. 

We used 
\num{99.99}\% 
of the cleaned \textit{Dergipark} documents for training and the rest for validation, to avoid over-inflation of the validation set due to the high number of documents. 
For \textit{YökTez}, 
\num{99.999}\%
of the documents were used for training. The final number of documents and the number of tokens after line and document-wise filtering of our scientific training corpora are listed in Table \ref{tab:train_data}. 
        
\subsection{Book Corpus}

The \textit{Book Corpus} is a compilation of \num{5080} Turkish fiction and non-fiction books. We cleaned the Book Corpus in a similar, but simpler heuristic. We first standardized the punctuation and removed invalid characters.  The initial \num{100} lines of each book have been filtered out if they contain author, translator, or publishing information. We dropped any line in the book that is all numeric or contains a URL or an e-mail. After the initial \num{70}\% lines, we truncated the lines after a keyword indicating a bibliography, notes, or a list of works of the author or the publishing house. \num{99.97}\% of the books were used for training (\num{5078} books), and the remaining two books for validation. 

\subsection{Bilkent Creative Writings}

The \textit{Bilkent Creative Writings} corpus comprises \num{8630} documents produced by Bilkent University students while taking creative writing courses in Turkish\footnote{\texttt{\href{https://github.com/selimfirat/bilkent-turkish-writings-dataset}{github.com/selimfirat/bilkent-turkish-\\writings-dataset}}}. We cleaned this data like the book corpus by removing lines containing special keywords (such as the word for assignment in Turkish) and truncating after bibliographies. 
\num{8457} of them were used for training and the rest was used for validation. 

\subsection{ParlaMintTR}

The \textit{ParlaMintTR} corpus is assembled from the CLARIN Flagship project\footnote{\texttt{\href{https://www.clarin.eu/parlamint}{clarin.eu/parlamint}}} and consists of the Turkish portion of parliamentary debates in Europe (\num{1335} documents). We used the original form of the debates without applying special cleaning or filtering. \num{1333} of the debates were used for training, and two debates for validation. 

\begin{table}
    \centering
    \resizebox{\linewidth}{!}{
    \begin{tabular}{llrr}
    \toprule

         \textbf{Corpus} & \textbf{Type} & \textbf{\# Docs} &  \textbf{\# Tokens (B)} \\ \midrule
       OSCAR \& mC4  & Web & \num{50336214}& \num{25.33}  \\  
       Dergipark  & Scientific & \num{334429} & \num{1.78}  \\
       Yöktez  & Scientific & \num{475817} & \num{15.24}  \\
       Books  & Literary & \num{5078} & \num{0.61}  \\
       Bilkent Creative Writings  & Creative Text & \num{8457} &  \num{0.01} \\
       ParlaMintTR  & Dialogue & \num{1333} &  \num{0.07} \\ \bottomrule
    \end{tabular}}
    \caption{Training Datasets}
    \label{tab:train_data}
\end{table}

\section{Methodology}

\subsection{Model}

We used an encoder-decoder Transformer model\footnote{Specifically, we used the version 1.1 of the official T5 implementation described at \texttt{\href{https://github.com/google-research/text-to-text-transfer-transformer/blob/main/released\_checkpoints.md\#t511}{github.com/google-research/text-to-text-transfer-\\transformer/blob/main/released\_checkpoints.md\#t511}}}~\cite{raffel2020exploring} for \turna. This choice was based on the finding that encoder-decoder models surpass decoder-only models when the UL2 objective is used, as demonstrated in \citealp{tay2023unifying}. Furthermore, the encoder component can still be employed effectively for understanding tasks when coupled with task-specific classification heads, thus reducing the model parameters by half. Due to our limited computational resources, we opted for the Large36L configuration~\cite{tay2021scale} for our model. This configuration requires only \num{37}\% of the parameters of a model configuration of comparable size, yet still outperforms it.

\turna\ has 36 encoder and decoder layers, each with 16 attention heads. The model's token embeddings are 1024 dimensional. The multi-layer perceptron layers have 2816 hidden dimensions and employ Gated GeLu activations~\cite{shazeer2020glu}.
The parameters of the input and classification layers are not shared.
These architectural choices result in a model with 1.1B parameters.
 
For tokenization, we used a unigram subword tokenizer~\cite{kudo-2018-subword} trained on 10GB of text that consists of random subsets of OSCAR~\cite{oscar_2022arXiv220106642A}, OPUS~\cite{opus_zhang2020improving} 
and Wikipedia dump dated September 17, 2021, using the SentencePiece implementation\footnote{\texttt{\href{https://github.com/google/sentencepiece}{github.com/google/sentencepiece}}}~\cite{kudo-richardson-2018-sentencepiece}.
This tokenizer\footnote{\texttt{\href{https://github.com/vngrs-ai/vnlp/tree/main/vnlp/turkish_word_embeddings}{github.com/vngrs-ai/vnlp/tree/main/\\vnlp/turkish\_word\_embeddings}}}
is provided by the VNGRS-AI Team. The initial vocabulary size of \num{32000} was expanded to \num{32128} with the addition of 128 sentinel tokens used by pretraining objectives.

\subsection{Pretraining Objectives}
The pretraining was performed with Mixture-of-Denoisers (MoD), consisting of several denoising objectives, which were shown to achieve better downstream performance~\cite{tay2023unifying}.
These objectives are R-denoising (regular denoising), S-denoising (sequential denoising), and X-denoising (extreme denoising), each characterized by 
the mean length of the corrupted spans,
the ratio of corrupted tokens,
and the number of corrupted spans.
R-denoising follows the standard span corruption method of T5, selecting spans of 2 to 5 tokens, covering about \num{15}\% of the input.
The task is then to predict the corrupted tokens in the decoder output.
S-denoising, on the other hand, corrupts a continuous portion from a random point in the input, accounting for approximately \num{25}\% of the input. 
Similar to R-denoising, this objective aims to predict a single corrupted span. However, it is similar to standard causal language modeling in the modeling approach.
X-denoising is designed as an interpolation between R-denoising and S-denoising. It aims to corrupt \num{50}\% of the input on average. This is achieved through a varying mix of many short or fewer long corrupted spans, exposing the model to both denoising and causal language modeling-like objectives.
During pretraining, these objectives are randomly assigned to each input sequence, with a distribution of \num{40}\% each for R- and X-denoisers and \num{20}\% for S-denoising.

The model differentiates between these denoisers by using specific sentinel tokens at the beginning of samples: \texttt{[NLG]} for X-denoiser, \texttt{[NLU]} for R-denoiser, and \texttt{[S2S]} for S-denoiser.

\subsection{Implementation details}

\paragraph{Pretraining.} We pretrained \turna\
for a total of \num{1740000} steps with a batch size of 48 and a source and target sequence length of 512 using a single v3-8 type TPU with the T5X\footnote{\texttt{\href{https://github.com/google-research/t5x}{github.com/google-research/t5x}}} library.
This configuration results in \turna\ being exposed to 42.7B tokens at the end of its training.
We disabled the dropout during pretraining but enabled it during fine-tuning.

The pretraining data is a mixture of samples from the collected datasets.
To ensure a fair representation of different language characteristics, we randomly selected samples from each dataset according to their proportions: Web Corpora (\num{50}\%), YökTez (\num{25}\%), DergiPark (\num{10}\%), Book Corpus (\num{10}\%), ParlaMintTR (\num{3}\%), and Bilkent Creative Writings (\num{2}\%).

\paragraph{Baselines.}

We compared our model with multilingual models: mT5, specifically mT5-large\footnote{\texttt{\href{https://huggingface.co/google/mt5-large}{hf.co/google/mt5-large}}}~\cite{xue-etal-2021-mt5}, and mBART\footnote{\texttt{\href{https://huggingface.co/facebook/mbart-large-cc25}{hf.co/facebook/mbart-large-cc25}}}~\cite{liu-etal-2020-multilingual-denoising}, as well as a monolingual encoder-only model, BERTurk\footnote{\texttt{\href{https://huggingface.co/dbmdz/bert-base-turkish-cased}{hf.co/dbmdz/bert-base-turkish-cased}}}~\cite{stefan_schweter_2020_3770924}, where applicable. 

\paragraph{Fine-tuning.}

We fine-tuned the models using Hugging Face's \texttt{transformers} library\footnote{\texttt{\href{https://github.com/huggingface/transformers}{github.com/huggingface/transformers}}}~\cite{wolf-etal-2020-transformers} on NVIDIA A40 GPUs. The standard text-to-text formulation is used for fine-tuning the encoder-decoder models, i.e., \turna, mT5 and mBART. Additionally, we fine-tuned \turna's encoder with a task-specific head for certain understanding tasks, referring to it as \turnaenc. 
The models were optimized for 10 epochs with an early stopping patience of 3 epochs.
We used the AdaFactor optimizer~\cite{shazeer2018adafactor} with a learning rate of \num{1e-3} to tune \turna\ and mT5 models, without a scheduler. 
However, our attempts at fine-tuning the mBART model with the AdaFactor optimizer did not yield a satisfactory training loss curve.
Consequently, we opted for the AdamW optimizer~\cite{loshchilov2017decoupled} with a learning rate of \num{5e-5} and a linear scheduler. The same optimizer and scheduler settings were applied for fine-tuning the BERTurk and \turnaenc\ models.
Due to our limited computational resources, we could not perform hyperparameter tuning and used the recommended fine-tuning settings for AdaFactor\footnote{\texttt{\href{https://huggingface.co/docs/transformers/main\_classes/optimizer\_schedules\#transformers.Adafactor}{hf.co/docs/transformers/main\_classes/\\optimizer\_schedules\#transformers.Adafactor}}} and default trainer settings\footnote{\texttt{\href{https://huggingface.co/docs/transformers/main\_classes/trainer\#trainer}{hf.co/docs/transformers/main\_classes/\\trainer\#trainer}}} for AdamW.
For each task and dataset, the batch size, and maximum input and target length parameters were individually selected, and their corresponding values can be found in Table~\ref{tab:finetuning-params}.  

We used beam decoding with a beam size of 4 and early stopping to generate predictions. For summarization and title generation tasks, we also applied a length penalty of 2 and enforced a no-repeat n-gram size of 3 to ensure the diversity of the output and prevent repetition of sequences.
\section{Experiments}

\subsection{Fine-tuning tasks}

This section provides an overview of downstream tasks used to evaluate the model. 
These tasks assess model capabilities across various domains, and include both natural language understanding and generation tasks. 
The understanding tasks include text classification, natural language inference, semantic textual similarity, named entity recognition, and part-of-speech tagging. 
The generation tasks comprise paraphrasing, summarization, and news title generation. 

\paragraph{Paraphrasing.} This task involves rephrasing a given text while retaining the original meaning. 
It assesses the model's understanding of semantics and its ability to generate diverse texts. 
We utilized two paraphrasing datasets, constructed from parallel corpora via machine translation and filtered based on semantic similarity~\cite{alkurdi-etal-2022-semantic}. 
These are TAT, which contains paraphrases from Tatoeba\footnote{\texttt{\href{https://tatoeba.org}{tatoeba.org}}}, and OST, which includes pairs from OpenSubtitles2018~\cite{lison-etal-2018-opensubtitles2018}.
\paragraph{Summarization.} Similar to paraphrasing, summarization also rephrases a text. 
However, it aims to produce a condensed version that only includes key information. 
Consequently, it imposes additional constraints on the model's generative capabilities. 
For evaluation, we used two datasets: TRNews~\cite{baykara2022abstractive}
and the Turkish subset of MLSUM~\cite{scialom-etal-2020-mlsum}.

\paragraph{News Title Generation.} Generating titles for news articles evaluates a model's ability to capture the most salient information in a concise manner and checks the model's creativity and understanding of key phrases in the news domain. 
We used the same two summarization datasets: TRNews and MLSUM. 

\paragraph{Named Entity Recognition.}

Named entity recognition (NER) aims to locate named entities, and subsequently classifies these entities into predefined categories, typically ``person'', ``location'' and ``organization''.
We employed two datasets for this task: WikiANN~\cite{wikiann-ner} and MilliyetNER~\cite{milliyet-ner}.

\paragraph{Part-of-speech Tagging.}
Part-of-speech (POS) tagging involves categorizing each word in a sentence according to its grammatical function. This task assigns a specific part of speech, such as noun, pronoun, or verb, to every word, clarifying its role within the sentence's structure. We used two Turkish Universal Dependencies~\cite{nivre-etal-2020-universal} treebanks, IMST~\cite{ud_turkish_imst-pos} and BOUN~\cite{ud_turkish_boun-pos}, to fine-tune and evaluate our model.

\paragraph{Semantic Textual Similarity.}

Semantic textual similarity (STS) tests the model's ability to contextually compare two sentences by producing a similarity score. We used the STSb-TR~\cite{beken-fikri-etal-2021-semantic} dataset to fine-tune and evaluate our model. 

\paragraph{Natural Language Inference.}
Natural language inference (NLI), also known as textual entailment, involves examining a pair of sentences, the premise and the hypothesis, to determine their relationship as ``entailment'', ``contradiction'', or ``neutral''. This task tests a model's understanding of context by comprehending the premise and assess if the hypothesis logically follows it. Therefore, NLI also measures a model's reasoning skills.
For this task, we used the Natural Language Inference in Turkish (NLI-TR) dataset~\cite{budur-etal-2020-data} for evaluation.  

\paragraph{Text Classification.}
Text classification involves categorizing texts into predefined groups based on their contents. 
This task assesses the model's contextual awareness and robustness in extracting relevant features from the input text, allowing it to discern important patterns and information crucial for accurate classification. 
We used three different datasets for evaluating this task: Product Reviews\footnote{\texttt{\href{https://huggingface.co/datasets/turkish_product_reviews}{hf.co/datasets/turkish\_product\_reviews}}}, TTC4900\footnote{\texttt{\href{https://www.kaggle.com/savasy/ttc4900}{kaggle.com/savasy/ttc4900}}}~\cite{yildirim2018ttc4900}, and Tweet Sentiments~\cite{amasyali2018words}.

\subsection{Evaluation Metrics}
We evaluated the generation tasks with ROUGE~\cite{lin-2004-rouge}, BLEU~\cite{Papineni02bleu:a} and METEOR~\cite{banarjee2005} metrics. For the understanding tasks, we adopted standard classification metrics such as accuracy, precision, recall, and F1. The only exception was semantic textual similarity, a regression task, for which we used the Pearson correlation coefficient for evaluation. For NLI and classification tasks, weighted precision, recall and F1 were reported, leaving out accuracy due to its equality to weighted recall. 

\subsection{Results}

\subsubsection{Generation Tasks}
We evaluated \textsc{Turna}'s generative capabilities on three tasks
and compared the results to mT5 and mBART. 
The results, as presented in Table~\ref{tab:generation}, show that \textsc{Turna} outperformed the baseline models in both paraphrasing and summarization, with mT5 ranking second and mBART last. 
In title generation, \textsc{Turna} performed the best on the TRNews dataset, followed by mBART. However, for the MLSUM dataset, mBART outperformed both \textsc{Turna} and mT5.

\begin{table}[H]
    \centering
    \caption{Downstream performance of models on generation tasks.}
    \label{tab:generation}
    \resizebox{\linewidth}{!}{
    \begin{tabular}{cllcccccc}
    \toprule
    \textbf{Task} & \textbf{Dataset} & \textbf{Model}  & \textbf{Rouge1} & \textbf{Rouge2} & \textbf{RougeL} & \textbf{BLEU} & \textbf{METEOR} \\ \midrule
    \multirow{6}{*}{\rot{Paraphrasing}} & \multirow{3}{*}{OST}      & mBART & \num{76.86} & \num{61.34} & \num{75.18} & \num{48.85} & \num{72.61} \\
                                 &                                  & mT5   & \num{77.49} & \num{62.15} & \num{75.87} & \num{49.66} & \num{73.61} \\ 
                                 &                                  & \textsc{TURNA} & \num{78.43} & \num{63.58} & \num{76.81} & \num{51.47} & \num{74.79} \\ \cmidrule{2-8}
                                 & \multirow{3}{*}{TAT}             & mBART & \num{82.77} & \num{68.68} & \num{81.31} & \num{55.57} & \num{77.34} \\
                                 &                                  & mT5   & \num{88.76} & \num{77.75} & \num{87.51} & \num{67.80} & \num{85.58} \\ 
                                 &                                  & \textsc{TURNA} & \num{90.22} & \num{80.23} & \num{88.95} & \num{71.14} & \num{87.56} \\ \midrule
    \multirow{6}{*}{\rot{Summarization}} & \multirow{3}{*}{MLSUM}   & mBART & \num{41.39} & \num{27.63} & \num{35.61} & \num{19.66} & \num{32.30} \\
                                         &                          & mT5   & \num{43.43} & \num{29.95} & \num{37.71} & \num{21.58} & \num{34.20} \\
                                         &                          & \textsc{TURNA} & \num{44.33} & \num{30.99} & \num{38.62} & \num{24.25} & \num{36.47} \\ \cmidrule{2-8}
                                         & \multirow{3}{*}{TRNews}  & mBART & \num{39.96} & \num{25.53} & \num{34.90} & \num{16.69} & \num{32.23} \\
                                         &                          & mT5   & \num{41.46} & \num{27.47} & \num{36.60} & \num{18.31} & \num{34.48}\\ 
                                         &                          & \textsc{TURNA} & \num{41.77} & \num{27.81} & \num{36.99} & \num{19.05} & \num{34.61} \\ \midrule
   \multirow{6}{*}{\rot{Title Generation}}  &   \multirow{3}{*}{MLSUM} & mBART & \num{32.97} & \num{19.71} & \num{31.32} & \num{7.41} & \num{18.29} \\
                                         &                          & mT5   & \num{32.60} & \num{19.65} & \num{30.93} & \num{7.15}  & \num{17.75} \\ 
                                         &                          & \textsc{TURNA} & \num{32.67} & \num{19.60} & \num{31.12} &  \num{7.08} & \num{17.90}\\ \cmidrule{2-8}
                                         & \multirow{3}{*}{TRNews}  & mBART & \num{35.40} & \num{21.92} & \num{34.32} & \num{11.95} & \num{23.26} \\
                                         &                          & mT5   & \num{34.84} & \num{21.62} & \num{33.85} & \num{11.96} & \num{22.40} \\
                                         &                          & \textsc{TURNA} & \num{36.47} & \num{22.88} & \num{35.47} & \num{12.64} & \num{23.62} \\ \bottomrule
    \end{tabular}}
    
\end{table}

\subsubsection{Understanding Tasks} \label{subsubsec:understanding}

In assessing understanding tasks, we compared both encoder-decoder models fine-tuned with the standard text-to-text formulation and encoder-only models, such as \textsc{Turna}-Encoder and BERTurk. \textsc{Turna} achieved results 
that surpass both mT5 and mBART across various tasks and datasets, as detailed in Tables~\ref{tab:pos-ner},~\ref{tab:nli}, and~\ref{tab:classification}, reporting POS tagging \& NER, NLI, and classification results, respectively. 
\turna\ outperformed mBART and mT5 in all classification, NLI, STS, POS tagging and NER tasks, except for the Milliyet (NER) dataset. 
While \turna\ slightly lagged behind BERTurk on some tasks, this was not surprising as encoder-decoder models often struggle with understanding tasks~\cite{lewis-etal-2020-bart, kementchedjhieva-chalkidis-2023-exploration}. 
However, \turnaenc\ surpassed \berturk\ in NER, NLI and some classification tasks, and was competitive in others.  The notable exception was the semantic textual similarity task (Table \ref{tab:sts}), where \turnaenc\ significantly lagged behind BERTurk. This suggests that further hyperparameter tuning could improve performance, as evidenced by an additional experiment where adjusting the learning rate enabled \turnaenc\ to achieve a
a significantly higher Pearson correlation score in the STS task (refer to Table \ref{tab:lr_sts} in the Appendix). 

\begin{table}[!ht]
 \centering
    \caption{Downstream performance of models on POS tagging and NER.}
    \label{tab:pos-ner}
    \resizebox{\linewidth}{!}{
    \begin{tabular}{@{}cllccccc@{}}
    \toprule
    \textbf{Task}                 & \textbf{Dataset} & \textbf{Model}  & \textbf{Precision} & \textbf{Recall} & \textbf{F1}    & \textbf{Accuracy} \\ \midrule
    \multirow{10}{*}{\rot{POS}} & \multirow{5}{*}{BOUN}  & mBART & \num{88.15}     & \num{87.75}  & \num{87.95} & \num{87.75}    \\
                               &                        & mT5   & \num{90.90}     & \num{90.74}  & \num{90.82} & \num{90.74}    \\

                               & & \textsc{TURNA} & \num{92.39}     & \num{92.35}  & \num{92.37} & \num{92.35} 
                               \\ \cmidrule{3-7}
                               & & BERTurk &    \num{90.60}  & \num{90.41}  & \num{90.50} &  \num{93.22}  \\
                               & & \textsc{TURNA}-Encoder & \num{90.30} & \num{90.31} &  \num{90.31} & \num{93.05} \\ \cmidrule{2-7}

                            & \multirow{5}{*}{IMST}     & mBART & \num{77.68}    & \num{77.40}  & \num{77.54} & \num{77.39}    \\
                            &                           & mT5   & \num{93.17}     & \num{93.05}  & \num{93.11} & \num{93.04}    \\
                            &                            & \textsc{TURNA} & \num{94.66}     & \num{94.48}  & \num{94.57} & \num{94.48}    \\ \cmidrule{3-7}
                             & & BERTurk &   \num{94.28}   &  \num{94.14} & \num{94.21} &  \num{95.62} \\
                             & & \textsc{TURNA}-Encoder & \num{93.34} & \num{93.27} & \num{93.31}  & \num{94.91} \\ \bottomrule

    \multirow{10}{*}{\rot{NER}} & \multirow{5}{*}{Milliyet}  & mBART &  \num{87.62}  & \num{70.67} & \num{78.23} & \num{98.11}    \\
                                 &                        & mT5   &  \num{84.73}   &  \num{71.98} & \num{77.83} & \num{98.20}    \\
                                 & & \textsc{TURNA} &  \num{91.36} & \num{83.28} & \num{87.13} & \num{97.91}    \\ \cmidrule{3-7}
                               & & BERTurk &  \num{93.51}    & \num{94.84}  & \num{94.17}  & \num{99.24}    \\
                               & & \textsc{TURNA}-Encoder &  \num{95.16}    & \num{96.03}  & \num{95.59}  & \num{99.46} \\ \cmidrule{2-7}

                              & \multirow{5}{*}{WikiANN}     & mBART & \num{90.76} & \num{89.12} & \num{89.93} & \num{95.84}    \\
                              &                           & mT5   &   \num{90.50}  & \num{89.90} & \num{90.20} & \num{95.93}    \\
                              &                            & \textsc{TURNA} & \num{90.48} & \num{90.20} & \num{90.34} & \num{96.18}    \\  \cmidrule{3-7}
                               & & BERTurk & \num{89.83} & \num{90.41} & \num{90.12} & \num{96.53} \\
                               & & \textsc{TURNA}-Encoder & \num{91.08} & \num{92.01} & \num{91.54} & \num{97.08} \\ \bottomrule
   \end{tabular}}
\end{table}

\begin{table}[!ht]
    \centering
    \caption{Downstream performance of models on natural language inference (NLI). 
    }
    \label{tab:nli}
    \resizebox{0.7\linewidth}{!}{
    \begin{tabular}{llcc}
    \toprule
     \textbf{Model}  & \textbf{Precision} & \textbf{Recall} & \textbf{F1}  \\ \midrule
     mBART          & \num{86.14} & \num{86.06} & \num{86.08}  \\
     mT5            & \num{83.67} & \num{83.66} & \num{83.66}  \\
     \textsc{TURNA}          & \num{86.20} & \num{86.19} & \num{86.19}  \\
                                 \midrule
    BERTurk         & \num{86.94} & \num{86.88} & \num{86.90}  \\
    \textsc{TURNA}-Encoder   & \num{88.28} & \num{88.30} & \num{88.28}  \\
                                \bottomrule
    \end{tabular}}

\end{table} 

\begin{table}[!ht]
 \centering
    \caption{Downstream performance of models on text classification.}
    \label{tab:classification}
    \resizebox{\linewidth}{!}{
    \begin{tabular}{@{}clccc@{}}
    \toprule
    \textbf{Dataset}                & \textbf{Model}  & \textbf{Precision} & \textbf{Recall} & \textbf{F1}    \\ \midrule
    \multirow{6}{*}{\rot{Product Reviews}} &  mBART            & \num{87.67}     & \num{93.63}  & \num{90.55}    \\ 
                                     & mT5              & \num{93.01}     & \num{94.17}  & \num{93.27}     \\
                                     & \textsc{TURNA}            & \num{94.67}     & \num{95.24}  & \num{94.81}     \\
                                     \cmidrule(l){2-5}
                                     & BERTurk          & \num{94.90}      & \num{95.44}  & \num{94.70}     \\ 
                                     & \textsc{TURNA}-Encoder      & \num{95.57} & \num{95.92}  & \num{95.67}    \\ \midrule
    \multirow{6}{*}{\rot{TTC4900}}   & mBART            & \num{78.23}     & \num{71.81}  & \num{73.08}     \\
                                     & mT5              & \num{67.52}     & \num{66.74}  & \num{66.80}     \\
                                     & \textsc{TURNA}            & \num{89.15}     & \num{88.11}  & \num{88.16}    \\
                                     \cmidrule(l){2-5}
                                     & BERTurk          & \num{91.97}     & \num{91.85}  & \num{91.88}     \\
                                     & \textsc{TURNA}-Encoder & \num{91.05}     & \num{90.53}  & \num{90.52}     \\\midrule
    \multirow{6}{*}{\rot{Tweet Sentiment}} & mBART            & \num{74.07}     & \num{71.85}  & \num{72.25}    \\ 
                                     &  mT5              & \num{68.20}      & \num{67.45}  & \num{66.71}     \\
                                     & \textsc{TURNA}            & \num{74.58}     & \num{73.78}  & \num{73.94}     \\ 
                                     \cmidrule(l){2-5}
                                     & BERTurk          & \num{75.91}  & \num{75.20}   & \num{74.79}      \\
                                     & \textsc{TURNA}-Encoder & \num{77.08}     & \num{76.82}  & \num{76.76}     \\ \bottomrule
\end{tabular}}
\end{table}

\begin{table}[H]
    \centering
    \caption{Downstream performance of models on semantic textual similarity (STS).
    }
    \label{tab:sts}
    \resizebox{0.45\linewidth}{!}{
    \begin{tabular}{llll}
    \midrule
    \textbf{Model}  &\textbf{Pearson}  \\ \midrule
    mBART     & \num{66.95}  \\
    mT5       & \num{59.40} \\
    \textsc{TURNA}     &  \num{78.74} \\ \midrule
    BERTurk     &  \num{82.60} \\
    \textsc{TURNA}-Encoder   &  \num{73.63}  \\ \bottomrule
    \end{tabular}
    }

\end{table}

\section{Conclusion}

In this study, we introduced \turna, a new Turkish language model that adopts an encoder-decoder architecture following the UL2 framework. 
This model was pretrained on a broad corpus covering web data, scientific articles, theses, books, creative writing, and parliament corpora. 
Our comprehensive evaluations across three generation and five understanding tasks on 13 different datasets showed that \turna\ outperforms existing multilingual models, mT5 and mBART, and performs better than or on par with the Turkish encoder-only model, \berturk. 
To encourage further research and facilitate benchmarking in Turkish NLP, we have made our models and the entire source code for data collection, filtering, model training, and fine-tuning publicly accessible. 
\section*{Limitations}

\turna, with its 1.1B parameters, excels in a variety of NLP tasks, surpassing similar-scale multilingual models like mT5 (1.2B) and mBART (610M) in both generation and understanding. 
However, its efficiency, especially in understanding tasks, is closely matched by the smaller, encoder-only model \berturk, which has only 110M parameters. This suggests that the scale-to-performance ratio of \turna\ may not be as efficient as expected. 

Addressing this, we modified \turna\ into \turnaenc\ by removing the decoder and adding task-specific heads, which improved its efficiency. \textsc{TURNA}-Encoder, having half the parameters of \turna, surpassed \berturk\ in most tasks, thereby improving its efficiency. However, the comparison with \berturk\ indicates a need for additional pretraining to fully leverage \turna's larger parameter count.

Current research on scaling laws indicates that training models for up to four epochs can be beneficial~\cite{taylor2022galactica, muennighoff2023scaling}. Despite having 1.1B parameters, \turna\ has been trained with approximately 43B tokens, which is roughly equivalent to one epoch. This under-training might be limiting its potential. Therefore, we suggest further pretraining of \turna\ to enhance its performance.

In our downstream evaluations, we used the same optimization hyperparameters across all tasks and datasets due to limited computational resources. This approach may have influenced performance as datasets carry differing sizes and tasks exhibit different difficulties. Hence, we suggest dataset and task-specific hyperparameter tuning to thoroughly demonstrate the capabilities of our model in downstream tasks.  


\section*{Acknowledgments}
We thank the Google TPU Research Cloud program for providing us with credits to pretrain our model on TPU v3-8 machines. 
We are grateful to TETAM and BOUN CMPE for providing access to the GPU cluster used in fine-tuning and evaluation experiments.
We also thank VNGRS-AI team for providing the tokenizer and the cleaned web corpus.
\bibliography{anthology,main}
\bibliographystyle{acl_natbib}

\appendix

\section{Appendix}
\label{sec:appendix}
\subsection{Cleaning Procedure for Scientific Corpus}
\label{sec:filtering}
Initially, we replaced invalid or misinterpreted characters resulting from Optical Character Recognition (OCR) errors, employing a predefined dictionary. Subsequently, we omitted preliminary text appearing before the abstract, which typically contains non-essential information such as affiliations and article metadata. This was achieved using regular expressions tailored for this purpose.
While this approach was sufficient for scientific articles, the theses posed additional challenges, including sections like lists of figures, tables, and customary declarations. To handle these sections, we relied on regular expressions designed to identify and subsequently discard specific titles and their accompanying content.

In our effort to maintain the quality of the extracted text from the PDF articles, we also implemented a line-wise filtering procedure involving the steps below:

\begin{itemize}
    \item \textbf{Text Statistics:} Each line from the articles was analyzed based on various statistics. These included character count, token count, numeric content, average token length, and metrics reflecting the prevalence of numbers, specifically the proportion of numeric tokens to total tokens and frequency of digit appearances. This stage ensured the removal of non-content elements, such as headers, page numbers, and table items.
    \item \textbf{Language Identification and Correction:} Given the potential presence of non-Turkish lines within the articles, each line was checked for its Turkish content using the \texttt{langid} library\footnote{\texttt{\href{https://github.com/saffsd/langid.py}{github.com/saffsd/langid.py}}}. In cases of potential anomalies or false detections, the surrounding lines were examined to correct such anomalies, ensuring that the majority of our extracted content is in Turkish.
    \item \textbf{Content Identification:} Although article metadata typically appears at the beginning of the documents, they may also appear elsewhere. To identify such elements as dates, email addresses, and names, each line was checked using specific regular expressions. Additionally, captions, identified by their distinct patterns, were detected and subsequently removed.
    \item \textbf{Identification and Filtering of Special Sections:} 
    In scientific texts, certain lines—like those in bibliographies and footnotes—may not contribute essential content, or they may even disrupt the primary narrative. To address this, we implemented strategies to detect and subsequently omit such lines. This step ensured the retention of the text's coherence and continuity.
    \item \textbf{Citation Filtering:} Citations, while crucial to academic papers, can interrupt text flow, especially when preparing data for language model training. We thus used patterns to identify and remove inline citations, guaranteeing a smooth textual flow.
\end{itemize}

After the line-wise filtering procedure was complete, we applied document-based filtering with the help of a Statistical Language Model (LM) trained on a compilation of May 2023 Turkish Wikipedia articles\footnote{\texttt{\href{https://huggingface.co/datasets/musabg/wikipedia-tr}{hf.co/datasets/musabg/wikipedia-tr}}}. A KenLM 5-gram language model was trained~\cite{heafield-2011-kenlm} on 6.8M sentences tokenized with a Turkish SentencePiece tokenizer\footnote{\texttt{\href{https://github.com/vngrs-ai/vnlp/tree/main/vnlp/turkish_word_embeddings}{github.com/vngrs-ai/vnlp/tree/main/\\vnlp/turkish\_word\_embeddings}}}. The KenLM model was then used to discard documents defined by separate thresholds for the \textit{Dergipark} articles (less than \num{5}\% LM score) and the \textit{YökTez} theses (less than \num{2}\% LM score). The thresholds have been selected by native Turkish speakers by analyzing the distribution of documents and their qualities based on document-based average LM score.

\subsection{Fine-tuning Datasets}
\paragraph{OST~\cite{alkurdi-etal-2022-semantic}} OST is a paraphrasing dataset, constructed by translating English subtitles from OpenSubtitles2018~\cite{lison-etal-2018-opensubtitles2018} into Turkish. The original subtitles and their translations were preprocessed to create an unfiltered version of the dataset with \num{1944955} pairs. These pairs were then filtered based on semantic similarity, resulting in a filtered version of the dataset with \num{706488} pairs.

\paragraph{TAT~\cite{alkurdi-etal-2022-semantic}} TAT is another paraphrasing dataset created using the same methodology as OST. The initial parallel corpus originates from Tatoeba\footnote{\texttt{\href{https://tatoeba.org}{tatoeba.org}}}. The unfiltered and filtered versions of the dataset include \num{265203} and \num{50423} pairs, respectively.

\paragraph{TR-News~\cite{baykara2022abstractive}} TR-News is a collection of news articles along with corresponding summaries and titles covering a wide range of topics. It is compiled from three Turkish national news outlets: Cumhuriyet, NTV, and HaberTürk. The dataset consists of approximately 307K articles, split into \num{277573} train, \num{14610} validation, and \num{15379} test documents.

\paragraph{MLSUM~\cite{scialom-etal-2020-mlsum}} MLSUM is a large-scale, multilingual summarization dataset that includes Turkish articles. The Turkish subset contains \num{273617} articles from InternetHaber, further divided into \num{259277} train, \num{11565} validation, and \num{12755} test documents.

\paragraph{WikiANN~\cite{wikiann-ner}} WikiANN is a multilingual named entity recognition dataset containing instances from Wikipedia articles annotated with tags of ``location'', ``person'', and ``organization''. The Turkish subset of the dataset includes \num{40000} rows, split into \num{20000} for training, \num{10000} for validation, and \num{10000} for testing.

\paragraph{MilliyetNER~\cite{milliyet-ner}} Milliyet NER is a named entity recognition dataset that includes instances from Turkish news articles annotated with tags of ``location'', ``person'', and ``organization''. The dataset comprises \num{515123} words, divided into a training set of \num{419996}, a validation set of \num{45532} and a test set of \num{49595} words. 

\paragraph{UD Turkish IMST~\cite{ud_turkish_imst-pos}} The IMST-UD Treebank is a Turkish dependency treebank in the format of the Universal Dependencies (UD) framework~\cite{sulubacak2018implementing}.
The treebank was annotated manually in a format other than UD, and then automatically converted for the UD version v1.3 to be the first Turkish UD treebank.
It has since then received various updates and corrections.
The latest version, v2.13, has \num{56422} tokens in total, with \num{36415} tokens for training, \num{10257} for validation, and \num{9750} for testing.

\paragraph{UD Turkish BOUN~\cite{ud_turkish_boun-pos}} The BOUN treebank is another Turkish dependency treebank that has been a part of the UD project since v2.7.
Since then, it has received a few updates with corrections.
The latest version, v2.13, has \num{121835} tokens in total, with \num{97797} tokens for training, \num{12023} for validation, and \num{12015} for testing.

\paragraph{STSb-TR~\cite{beken-fikri-etal-2021-semantic}} STSb-TR is derived from the English Semantic Textual
Similarity benchmark (STSb) dataset~\cite{cer-etal-2017-semeval} by translating the English sentences into Turkish using Google Translate, with no manual corrections. Each data element has two sentences and a corresponding similarity score. The dataset contains \num{5749} training, \num{1500} validation and \num{1379} test samples. 
\paragraph{NLI-TR~\cite{budur-etal-2020-data}} The Natural Language Inference in Turkish (NLI-TR) dataset consists of two large-scale datasets containing pairs of sentences labeled as ``entailment'', ``contradiction'', or ``neutral''. These sentence pairs were obtained by translating the widely used NLI corpora, made up of SNLI~\cite{snli:emnlp2015} and MultiNLI~\cite{N18-1101}. The SNLI dataset includes 570K samples, with 550K for training, 10K for validation, and 10K for testing. The MultiNLI dataset contains 413K samples, with 393K for training and 20K for validation, evenly divided between matched and mismatched pairs. 

\paragraph{Product Reviews}

The Turkish Product Reviews is a sentiment classification dataset that contains product reviews from various online sources, and is available on Hugging Face\footnote{\texttt{\href{https://huggingface.co/datasets/turkish_product_reviews}{hf.co/datasets/turkish\_product\_reviews}}}. A total of \num{235165} reviews are categorized as positive or negative. We deduplicated the dataset before usage, and split it with an 80-10-10 train-validation-test ratio. The resulting dataset contains \num{186806} training, \num{23351} validation and \num{23351} test samples. 

\paragraph{TTC4900~\cite{yildirim2018ttc4900}} 

The dataset is made available by the Kemik NLP Group\footnote{\texttt{\href{http://www.kemik.yildiz.edu.tr}{kemik.yildiz.edu.tr}}}, and contains \num{4900} news articles and texts classified with one of seven categories: economy, culture-arts, health, politics, sports, technology and world. The dataset is available on Kaggle\footnote{\texttt{\href{https://www.kaggle.com/savasy/ttc4900}{kaggle.com/savasy/ttc4900}}} and Hugging Face\footnote{\texttt{\href{https://huggingface.co/datasets/ttc4900}{hf.co/datasets/ttc4900}}}. The TTC4900 data was also deduplicated before fine-tuning, and split with an 80-10-10 ratio, leaving \num{3631} samples for training, and \num{454} samples each for test and validation.

\paragraph{Tweet Sentiments~\cite{amasyali2018words}}

Tweet Sentiments is a sentiment classification dataset with three categories: positive, negative and neutral. The dataset consists of \num{17289} tweets that contain comments about a GSM operator, split into \num{13832} training and \num{3457} test samples. Due to lack of a validation set, the training set was split with a 90-10 train-validation ratio. After deduplication, the resulting fine-tuning dataset contains \num{12421} training, \num{1381} validation and \num{3456} test samples. 

\subsection{Fine-tuning details}
\paragraph{Data splits.} We used predefined splits for datasets, including training, validation, and test sets. For datasets lacking both validation and test sets, we divided the data into training, validation, and test sets with an 80-10-10 ratio. In the absence of the validation set only, we utilized \num{10}\% of the original training data to generate a validation set, while the remaining \num{90}\% was used for training. We used the same approach for datasets that lacked a test set. For the NLI task, we fine-tuned our model on the training set referred to as NLI-TR~\cite{budur-etal-2020-data}, which is the combination of the training sets of SNLI-TR and MultiNLI-TR, and we used the already existing test and validation sets of the SNLI-TR dataset.

\paragraph{Dataset-specific parameters.}  Considering the varying lengths of dataset samples, we used dataset-specific parameters. These parameters set the maximum input and target lengths, and batch size to fit into the largest batch. In order to speed up the fine-tuning process, we employed bf16 mixed precision in the summarization and title generation experiments, allowing for a larger batch size.
Table~\ref{tab:finetuning-params} shows the hyperparameters used for fine-tuning.

\begin{table*}
\caption{\label{tab:finetuning-params}Dataset-specific hyperparameters for fine-tuning}
\begin{tabular}{llccc}
\toprule
\textbf{Task}                     & \textbf{Dataset}    & \textbf{Max Input Length} & \textbf{Max Target Length} & \textbf{Batch Size} \\ \midrule
\multirow{2}{*}{Summarization}    & TRNews              & \num{768}                       & \num{128}                        & \num{4}                     \\
                                  & MLSUM               & \num{768}                       & \num{128}                        & \num{4}                     \\ \midrule
\multirow{2}{*}{Title Generation} & TRNews              & \num{256}                       & \num{64}                         & \num{8}                     \\
                                  & MLSUM               & \num{256}                       & \num{64}                         & \num{8}                     \\ \midrule
\multirow{2}{*}{Paraphrasing}     & Tatoeba             & \num{20}                        & \num{20}                         & \num{128}                   \\
                                  & OpenSubtitles       & \num{20}                        & \num{20}                         & \num{128}                   \\ \midrule 
\multirow{2}{*}{NER}              & WikiANN             & \num{60}                        & \num{40}                         & \num{64}                    \\
                                  & MilliyetNER         & \num{380}                       & \num{60}                         & \num{8}                     \\ \midrule 
\multirow{2}{*}{POS}              & BOUN                & \num{90}                        & \num{300}                        & \num{8}                     \\
                                  & IMST                & \num{60}                        & \num{210}                        & \num{16}                    \\ \midrule 
NLI                               & NLI-TR              & \num{128}                       & \num{8}                          & \num{32}                     \\ \midrule
\multirow{3}{*}{Classification}   & Product Reviews     & \num{20}                        & \num{4}                          & \num{32}                    \\ 
                                  & TTC4900             & \num{1450}                      & \num{8}                          & \num{2}                     \\
                                  & Tweet Sentiment     & \num{160}                       & \num{4}                          & \num{32}                     \\ \midrule 
STS                               & STSb-TR             & \num{140}                       & \num{10}                         & \num{32}                    \\ \midrule
                                  
\end{tabular}
\end{table*}

\subsection{Mode-Switching}

In the UL2 framework, specific sentinel tokens are dedicated to different pretraining objectives, enabling the model to adjust its mode for optimal task performance. This approach is also applied to fine-tuning and few-shot learning by using a token tailored to the needs of the downstream task, such as \texttt{[S2S]} for generation tasks. This is known as mode switching.

We tested mode switching by fine-tuning \textsc{TURNA} on several tasks and datasets. The results, detailed in Tables~\ref{tab:mode_paraphrasing},~\ref{tab:mode_classification}, and~\ref{tab:mode_sts}, showed that \textsc{TURNA} models fine-tuned without any sentinel token scored highest on paraphrasing evaluations. However, a separate sentinel token achieved the best scores on different classification datasets, with the scores being remarkably close. In the semantic textual similarity task, the model trained with the \texttt{[NLG]} token performed the best.

We found no consistent pattern in the performance of different tokens across various tasks and datasets. This suggests that mode-switching might not always enhance performance, and could potentially degrade it.

\begin{table}[!ht]
    \centering
    \caption{Comparison of mode switching modes on the paraphrasing task.}
    \label{tab:mode_paraphrasing}
    \resizebox{\linewidth}{!}{
    \begin{tabular}{ccccccc}
    \toprule
    \textbf{Dataset}& \textbf{Mode}& \textbf{Rouge1}& \textbf{Rouge2}& \textbf{RougeL}& \textbf{BLEU}& \textbf{METEOR}\\ \midrule
    \multirow{4}{*}{OST}              &  \texttt{-}                              & \num{78.43}                            & \num{63.58}                            & \num{76.81}                            & \num{51.47}                           & \num{74.79}                            \\
                                      & \texttt{[NLG]}                            & \num{76.20}                            & \num{61.11}                            & \num{74.50}                            & \num{46.27}                          & \num{73.76}                            \\
                                      & \texttt{[NLU]}                            & \num{77.18}                            & \num{61.97}                            & \num{75.33}                            & \num{48.39}                          & \num{74.02}                            \\
                                      & \texttt{[S2S]}                            & \num{77.20}                            & \num{61.98}                            & \num{75.44}                            & \num{48.53}                          & \num{74.05}                            \\ \midrule
    \multirow{4}{*}{TAT}              &  \texttt{-}                              & \num{90.22}                            & \num{80.23}                            & \num{88.95}                            & \num{71.14}                           & \num{87.56}                            \\
                                      & \texttt{[NLG]}                            & \num{89.66}                            & \num{79.28}                            & \num{88.41}                            & \num{69.54}                          & \num{87.18}                            \\
                                      & \texttt{[NLU]}                            & \num{89.08}                            & \num{78.53}                            & \num{87.90}                            & \num{68.33}                          & \num{86.82}                            \\
                                      & \texttt{[S2S]}                            & \num{89.71}                            & \num{79.37}                            & \num{88.45}                            & \num{69.61}                          & \num{87.26}                            \\ \bottomrule
\end{tabular}}
\end{table}

\begin{table}[H]
    \centering
    \caption{Comparison of mode switching modes on the text classification task.}
    \label{tab:mode_classification}
    \resizebox{\linewidth}{!}{
    \begin{tabular}{ccccccc}
    \toprule
    \textbf{Dataset}  & \textbf{Mode}  &\textbf{Precision} & \textbf{Recall} & \textbf{F1}  \\ \midrule
    \multirow{4}{*}{Product Reviews}        & \texttt{-}   & \num{94.67} & \num{95.24} & \num{94.81}  \\
                                            & \texttt{[NLG]} & \num{94.30} & \num{95.03} & \num{94.39}  \\
                                            & \texttt{[NLU]} & \num{94.45} & \num{95.10} & \num{94.60}  \\
                                            & \texttt{[S2S]} & \num{94.34} & \num{95.04} & \num{94.47}  \\
                                \midrule
                    
    \multirow{4}{*}{TTC4900}     & \texttt{-}   & \num{89.15} & \num{88.11} & \num{88.16}  \\
                                 & \texttt{[NLG]} &  \num{89.50} & \num{88.33} & \num{88.39}    \\
                                 & \texttt{[NLU]} &  \num{86.18} & \num{84.14} & \num{84.31}\\
                                 & \texttt{[S2S]} & \num{90.83} & \num{90.31} & \num{90.24}    \\

                                \midrule
    \multirow{4}{*}{Tweet Sentiment}       
                                 & \texttt{-} & \num{74.58} & \num{73.78} & \num{73.94}  \\
                                 & \texttt{[NLG]} & \num{76.01} & \num{75.84} & \num{75.56}   \\
                                 & \texttt{[NLU]} &  \num{75.45} & \num{75.46} & \num{75.45} \\
                                 & \texttt{[S2S]}  & \num{75.55} & \num{74.91} & \num{74.86}    \\
    
                                \bottomrule
    \end{tabular}}

\end{table}

\begin{table}[H]
    \centering
    \caption{Comparison of mode switching modes on semantic textual similarity (STS).}
    \label{tab:mode_sts}
    \resizebox{0.4\linewidth}{!}{
    \begin{tabular}{ccc}
    \textbf{Mode}  &\textbf{Pearson}  \\ \midrule
    \texttt{-}    &  \num{78.74} \\
     \texttt{[NLG]} & \num{79.71} \\
     \texttt{[NLU]} & \num{78.45} \\
     \texttt{[S2S]} & \num{78.30} \\ \bottomrule
    \end{tabular}} 
\end{table}

\subsection{Hyperparameter Tuning}

We conducted an additional experiment on the Semantic Textual Similarity task due to the low Pearson correlation score obtained by \textsc{TURNA}-Encoder when compared to \textsc{TURNA} and BERTurk. We fine-tuned \textsc{TURNA}-Encoder with different learning rates on the regression task. The results are reported in Table \ref{tab:lr_sts}. The difference in Pearson correlation scores suggest that elaborate hyperparameter tuning can significantly alter the downstream performance of our model. 

\begin{table}[H]
    \centering
    \caption{Comparison of \textsc{TURNA}-Encoder performance with different learning rates on semantic textual similarity (STS).}
    \label{tab:lr_sts}
    \resizebox{0.7\linewidth}{!}{
    \begin{tabular}{rcc}
    \textbf{Learning Rate}  &\textbf{Pearson}  \\ \midrule
    (Default) \num{5e-5}    & \num{73.63}  \\
     \num{5e-4} & \num{77.13} \\
     \num{5e-3} & \num{-3.56} \\
     \num{5e-2} & \num{17.92} \\ \bottomrule
    \end{tabular}} 
\end{table}

\end{document}